# Deep Convolutional Neural Network Applied to Electroencephalography: Raw Data vs Spectral Features

Dung Truong, Michael Milham, Scott Makeig, Arnaud Delorme

*Abstract—* The success of deep learning in computer vision has inspired the scientific community to explore new analysis methods. Within the field of neuroscience, specifically in electrophysiological neuroimaging, researchers are starting to explore leveraging deep learning to make predictions on their data without extensive feature engineering. This paper compares deep learning using minimally processed EEG raw data versus deep learning using EEG spectral features using two different deep convolutional neural architectures. One of them from Putten et al. (2018) is tailored to process raw data; the other was derived from the VGG16 vision network (Simonyan and Zisserman, 2015) which is designed to process EEG spectral features. We apply them to classify sex on 24-channel EEG from a large corpus of 1,574 participants. Not only do we improve on state-of-the-art classification performance for this type of classification problem, but we also show that in all cases, raw data classification leads to superior performance as compared to spectral EEG features. Interestingly we show that the neural network tailored to process EEG spectral features has increased performance when applied to raw data classification. Our approach suggests that the same convolutional networks used to process EEG spectral features yield superior performance when applied to EEG raw data.

## I. Introduction

Electroencephalography (EEG) is a neuroimaging technique that allows measuring brain activities at the speed of thoughts and action, in both typical laboratory settings and natural environments. EEG is widely used in cognitive neuroscience, clinical research, and brain-computer interface - communication channels that bypass the natural output pathways of the brain - to allow brain activity to be directly translated into directives that affect the user's environment.

Deep learning (DL) is a powerful tool that can extract abstract patterns from complex digital signals without much (if any) feature engineering and produce impressive classification results in various fields such as natural language processing and computer vision [1]. Hand-engineered features are still commonly used in DL-EEG research. A recent literature review surveying all recent DL-EEG publications [1] found that 49% of papers used hand-engineered features, from which 38% corresponded to frequency domain-derived features. We believe that there is tremendous potential in applying DL directly on minimally processed raw EEG data, both in terms of performance in various tasks and generalization ability. DL-EEG also has potential as a tool for knowledge discovery and automatic hypothesis generation. Van Putten et al. [2] showed that Convolutional Neural Network (CNN) may be applied to raw EEG data during relaxation periods to predict participants' sex with more than 80 percent accuracy (age 18–98; 47% males).

This project aims to replicate [2] on a larger dataset and to compare its performance applied to raw EEG data versus EEG spectral features. The dataset, collected and made publicly available by the Child Mind Institute Healthy Brain Network project, contains resting EEG data from more than a thousand juvenile (5-22 years) participants [3]. We also aim to assess the performance of a DL model specifically tailored to process EEG spectral data. To this aim, we repurposed the VGG-16 model originally applied to EEG spectral data [4] and applied both models to both raw and spectral data.

Henceforth, we call our application of the CNN model described in [2] to raw data R-SCNN (Raw/Sex CNN). R-SCNN repurposed for spectral data is called S-SCNN. Similarly, we refer to our modified VGG-16 model applied to raw EEG data as R-VGG and use S-VGG for the version of the model trained on spectral data.

## II. Methods

**EEG recordings**. High-density EEG data were recorded in a sound-shielded room at a sampling rate of 500 Hz with a bandpass of 0.1 to 100 Hz, using a 128-channel EEG geodesic hydrogel system by Electrical Geodesics Inc. (EGI) [3]. The data are publicly available for download at http://fcon_1000.projects.nitrc.org/indi/cmi_healthy_brain_network. We only considered the resting data files. These were 6 minutes in length and were composed of successive 20-s to 40-s periods of eyes open and eyes closed rest respectively.

**Raw data preprocessing**. Although DL may be applied to raw EEG data without any preprocessing [1], we minimally preprocessed the data following the practice in [2] using EEGLAB v2021 [5] running on MATLAB 2020b. We used only eye-closed data segments (~170s per subject), ignoring the first and last 3 seconds of each eye-closed period (resulting in five periods of 34 seconds). We removed the mean baseline for each data epoch from each channel, down-sampled the data to 128 Hz, and subsequently band-pass filtered the data between 0.25–25 Hz (FIR filter of order 6601; 0.125 Hz and 25.125 Hz cutoff frequencies (-6 dB); zero phase, non causal). Data were re-referenced to the averaged mastoids and cleaned using Artifact Subspace Reconstruction EEGLAB plug-in *clean_rawdata* (v2.3) [6], an automated method that removes artifact-dominated channels and portions of data (parameters used were 5 for FlatLineCriterion, 0.7 for ChannelCriterion, and 4 for LineNoiseCriterion). Removed channels were then interpolated using 3-D spline interpolation (EEGLAB *interp.m* function). No bad portions of data were removed. While our recordings have 128 channels data, the comparison study [2] used only 24 channels. From the 128 available channels, we thus selected, by visual inspection of the overlaid channel maps, the 24 closest channels to the montage used in [2]. Finally, we segmented eye-closed data periods into non-overlapping 2-s windows: each preprocessed 2-s epoch was used as a sample for our final dataset. Each subject provided about 81 2-s samples (mean 80.8 ± 3.32). Each sample in our dataset thus had dimension 24x256 (24 channels and 2(s) x 128(Hz) time points) (Fig. 1a). The 128-Hz down sampling and 2-s window length were identical to those used in [2]. No bad epochs were removed. No further preprocessing was performed for learning from the raw data.

**EEG-PSD images-based features.** Using the preprocessed data above, for each EEG channel we extracted traditional power spectral density (PSD) scalp maps in three EEG bands, theta (4-7 Hz), alpha (7-13 Hz), and beta (13-25 Hz). These EEG bands were chosen to match [4]. In [4], the beta frequency band extended to 30 Hz, although it was capped here at 25 Hz as the raw data were low pass filtered below 25 Hz. PSD was calculated using the Welch method by averaging PSDs derived from the FFT of 39% overlapping 0.5-s hamming-tapered windows. We then used the computed EEG-PSD features to plot power spectrum heat maps for the three EEG bands using bicubic 2-D interpolation (EEGLAB *topoplot.m* function). Image pixel values were rescaled to be between 0 and 255 for each channel and values outside the disk outlining the head limit were set to 0. These images contain the topographical information about scalp signal power in the three frequency bands of interest. The three scalp topographies may be combined into a chromatic image (Fig. 1b) for the S-VGG model or placed side by side to form a 2-D image for the S-SCNN model (Fig. 1c).

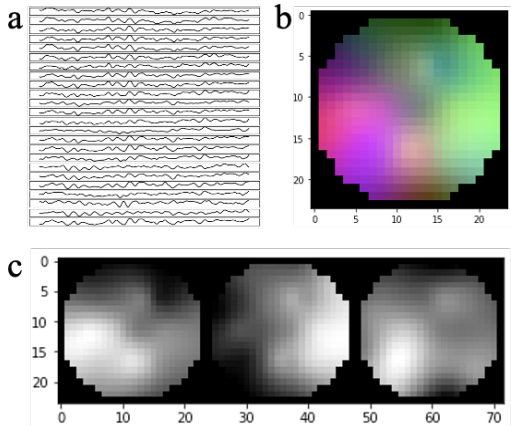

**Figure 1**. Raw and spectral data samples. (a) shows a 24x256 raw EEG sample. (b) shows a combined 24x24x3 scalp topography spectral data. (c) shows a 24x72 scalp topography of the three frequency bands represented side-by-side.

**R-SCNN.** This network used the 2-D 24x256 raw data matrices as input (Fig. 1a). We attempted to reconstruct the network architecture from the original paper [2] but faced challenges as the details necessary to replicate the network were in some cases missing and in others inconsistent. Our best effort at replication is summarized in Table 1. Each of the first 4 CNN layers is followed by a max pooling layer then a dropout layer using a 25% dropout rate. The output of each convolutional and fully connected (FC) layer (except the last) was transformed by a Rectified linear unit (ReLU). The classification layer is a 2-unit FC layer with a softmax activation, resulting in a probability $p$ for male or female sex ($p < 0.5$ for males and $p \geq 0.5$ for female). The number of network trainable parameters was 12,713,934.

| Layer | Filter size | # of filters/hidden units |
|---|---|---|
| Convolutional | 3x3 | 100 |
| MaxPooling Dropout (25%) | | |
| Convolutional | 3x3 | 100 |
| MaxPooling Dropout (25%) | | |
| Convolutional | 2x3 | 300 |
| MaxPooling Dropout (25%) | | |
| Convolutional[†] | 1x7 | 300 |
| MaxPooling[*] Dropout (25%) | | |
| Convolutional[†] | 1x3 | 100 |
| Convolutional[†] | 1x3 | 100 |
| Fully connected | | 6144 |
| Fully connected | | 2 |
| Softmax | | |

**Table 1. R-SCNN and S-SCNN configurations.** The ReLU activation function is not shown for brevity. All convolutional layers have stride 1 and no padding except for S-SCNN where padding is 1 in layers indicated by [†]. All pooling layers have window size 2x2, stride 2, and no padding, except for the last layer (indicated by [*]) in S-SCNN where window size is 1x2 and stride 1.

**S-SCNN.** This network used 2-D spectral topography matrices of size 24x72 as input (Fig. 1c) which allowed reuse of the R-CNN architecture, albeit with decreased input size. To accommodate for the reduction in input size, we zero-padded the final three convolutional layers (see Table 1), while the rest of the network remained unchanged. With this configuration, the output of the last convolutional layer was flattened to a 1-D vector of size 1400, compared to 1900 in the R-SCNN. In total, this model comprised 9,641,934 trainable parameters.

| Layer | Filter size | # of filters/hidden units |
|---|---|---|
| Convolutional[†] | 3x3 | 16 |
| Convolutional | 3x3 | 16 |
| MaxPooling | | |
| Convolutional | 3x3 | 32 |
| Convolutional | 3x3 | 32 |
| MaxPooling | | |
| Convolutional | 3x3 | 64 |
| Convolutional | 3x3 | 64 |
| Convolutional | 3x3 | 64 |
| MaxPooling | | |
| Fully connected | | 1024 |
| Dropout (50%) | | |
| Fully connected | | 1024 |
| Dropout (50%) | | |
| Fully connected | | 2 |
| Softmax | | |

**Table 2. S-VGG and R-VGG configurations.** The ReLU activation function is not shown for brevity. All convolutional layers have stride 1 and padding 1. The first convolutional layer (indicated by [†]) accepts different input sizes for the two models. All pooling layers have window size 2x2, stride 2, and no padding.

**S-VGG.** VGG-16 was originally designed to be trained on 15M images of dimension 256x256x3 [4]. A previous publication reported success in applying a vision-specific model such as VGG-16 to EEG spectral data [4]. Our scalp images had dimension 24x24x3 (Fig. 1b). Since the input size was smaller than in [4], the number of convolutional layers was reduced accordingly by omitting layers 19-32 of VGG-16. We also divided the number of filters and hidden units in the convolutional and FC layers by 4 to reflect our lower number of training samples (Table 2). In total, this model contained 1,751,506 trainable parameters.

**R-VGG.** To adapt S-VGG to the 24x256 raw data matrices (Fig. 1a), we decreased the number of input channels in the first convolutional layer from 3 to 1. The rest of the network remained unchanged. In total, this model contained 7,452,850 trainable parameters.

### III. SIMULATIONS

**Training, validation, and test sets.** While the dataset we used includes 2,224 participants, there were only 787 females (35%). We decided to use 1574 participants (50% female) to ensure class balance, by selecting the first 787 male from a list of participants ordered by their IDs. Following [2], we then split the balanced data into training, validation, and test sets in size ratio 60:30:10. Each segment received a binary label, indicating a male (0) or a female (1). This gave 71,300 samples (885 participants; 49.94% female) for training, 39,868 samples (492 subjects; 50% female) for validation, and 16,006 samples (197 subjects; 50.3% female) for testing.

**Experimental setup.** All models were trained on a single NVIDIA V100 SMX2 GPU (32 GB) with Python 3.7.10 and PyTorch 1.3.1. During training, the validation data were used to assess its performance and to inform a stopping rule. We trained all four models using an Adamax optimizer with default hyperparameters (learning rate = 0.002, $\beta_1$ = 0.9, $\beta_2$ = 0.999, $\epsilon$ = 1e-08) except for setting decay = 0.001. Batch size was set at 70, following [2]; training was performed for 70 epochs. No other hyperparameter tuning nor batch normalization was performed.

**Statistics.** To assess the robustness of our classification results, we trained each model 10 times from different random seeds giving 10 different weight initializations, allowing us to calculate the 95% confidence interval for mean performance (± [standard deviation / sqrt(10)]*1.96).

**Early stopping.** Overfitting is a common issue in deep learning. As shown in Fig. 2, all four models overfit the training data. One common DL practice to avoid overfitting is early stopping, in which training is stopped (and model performance evaluated) when validation accuracy starts to plateau or decrease as training accuracy continues to grow [7]. We saved intermediary models during training and since each of the models overfit the data at different rates, we chose to evaluate the models at different training epochs (Fig. 2). We noticed that both VGG models overfit the data faster than R-SCNN and S-SCNN. When compared between raw vs spectral models, validation accuracy of both models trained on spectral data plateaued within the first 10 training epochs; that of S-VGG stopped increasing after only 1 epoch. For raw data, we assessed performance after 20 epochs; for the spectral data, we assessed performance after 10 epochs.

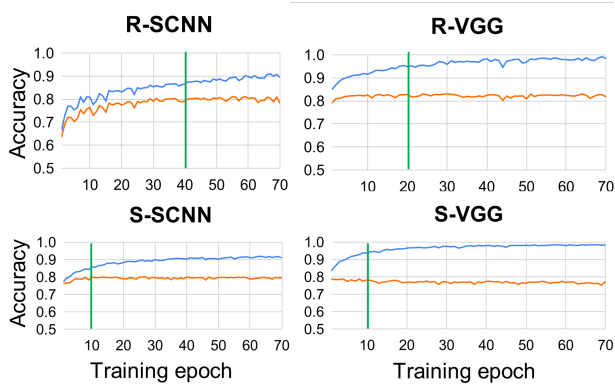

**Figure 2**. Training (blue) and validation (orange) curves for each of the models for a particular random seed. Vertical green lines indicate the epochs at which the models performance was assessed.

### IV. RESULTS

**Evaluation metric.** Per-sample prediction accuracy was reported for all models. Following [2], we also obtained a final performance estimate for the test dataset by taking the mean gender probability $p_{ave}$ of the first 40 2-s samples for each subject; if $p_{ave} > 0.5$, the subject was classified as 1 (female) or as 0 (male) otherwise. We refer to this as per-subject performance. Table 3 below shows per-sample and per-subject results for all of our four models.

| Model | Per-sample | Per-subject |
|---|---|---|
| R-SCNN | 80.6 (79.7 to 81.5) | 85.1 (84.3 to 85.9) |
| R-VGG | **83.1** (82.7 to 83.4) | **87.0** (86.6 to 87.4) |
| S-SCNN | 79.0 (78.7 to 79.3) | 83.2 (82.1 to 84.3) |
| S-VGG | 77.1 (76.8 to 77.4) | 81.3 (80.0 to 82.6) |

**Table 3.** Models classification accuracy. 95% confidence interval is indicated in parenthesis. Bolded values indicate best performance.

**Performance on raw EEG.** Only per-subject classification result (using votes on the first 40 data segments) was reported in [2], achieving 81% prediction accuracy. Here we report both per-sample and per-subject classification performance and show improvement in prediction accuracy for both models trained on the raw EEG data. R-SCNN achieved 85.1% accuracy, while R-VGG achieved 87% per-subject performance. Because there is no overlap between the 95% confidence intervals (equivalent to an unpaired parametric t-test), the difference is statistically significant at the p=0.05 threshold.

**Performance on spectral EEG.** S-VGG performed significantly worse than S-SCNN on per-sample performance. R-SCNN significantly (about 1%) outperforms S-SCNN on per-sample performance. In all cases, R-VGG gave significantly better performance.

### V. DISCUSSION

We have shown that a neural network tailored to process simple EEG spectral features gave improved (sex) classification performance when applied instead to the raw data. This model also outperformed a previous sex classification deep learning approach [2]. Our approach suggests that the same convolutional networks used to process image-based inputs, specifically scalp images of EEG spectral features, can give superior performance when applied instead to EEG raw data. Despite the popularity of reusing vision inspired convolutional neural network architecture [1] to process spectral scalp topographies, such preprocessing is not warranted.

**Reproducibility issue.** We faced many challenges replicating the network architecture in [2]. Neither the code nor the data used in the original paper [2] were published, hindering our replication attempt. We suggest that all deep learning experimental papers be accompanied by the code and data used unless there is some data privacy restriction. Here we used a publicly available dataset, detailed our choices of all necessary parameters, and provided all scripts in a documented GitHub repository: https://github.com/dungscout96/DL-EEG.

**CNN applied to time series.** It is not common to apply CNN to biophysical time series [1]. In general, recurrent neural networks are used for this type of applications (such as networks with LSTM or GRU units [1]). These architectures often require considerable training time because they are, by construction, iterative and more difficult to parallelize. However, in most applications where time is involved such as language or biological time series, bidirectional recurrent neural networks are used. CNN processes the time dimension as space so they are bi-directional by definition. In the future, we intend to compare in more details the pros and cons of using recurrent neural networks vs CNNs on EEG time series.

**Limitations of spectral approach with vision-inspired CNNs.** As we only used three frequency bands [4], one might argue that we did not use enough or the most optimal bands. Using more frequency bands would prevent the use of standard trichromatic vision-inspired CNNs. One may also argue that the spectral models have fewer trainable parameters than those of the raw data models, explaining the inferior performance. However, as seen in Fig. 2, the spectral models heavily suffered from overfitting, indicating that the models might already be too expressive for the task. At this point, our work should be seen as a demonstration that applying vision-inspired CNNs to basic spectral decomposition is limited and need not be preferred over minimally preprocessed raw data by default. Only using both the amplitude and phase information at all frequencies may the spectral data be used to reconstruct the raw data. In all likelihood CNNs applied to this type of spectral data should yield similar performance as when using raw data since there is no information loss.

**Known sex differences affecting EEG signals.** Girls have generally thinner skulls than boys at the same age, allowing more of the cortical field signals to reach the scalp [8]. It would be of interest to train our networks on longitudinal per-subject normalized data to determine how large a role this difference may play in our results. Adding participant age as an input might also increase network performance.

**Raw data 2-D channel x time data organization.** When EEG channels are organized in a 1-D vector, depending on the chosen channel order, neighboring channels on the scalp may not be contiguous in the 1-D vector. As a consequence, the first convolutional layers of CNNs might not be able to extract local scalp features optimally. Ideally, raw data should be organized as a 3-D vector with interpolated scalp topographies for each time latency. We intend to explore this research direction in the future.

**Amount of data.** Our R-SCNN network (designed to follow [2]) gave better performance than previously reported in [2]. This could be because of our larger dataset or difference in age range. The larger amount of data we used might also explain the superiority of the networks using raw data as input compared to those using the frequency reduced spectral data. At this point, we can only claim that given enough data, approaches using raw data as input may yield superior results. In the future, it might be also possible to apply our pretrained raw data CNN architectures to new problems with smaller data amounts by only retraining the last layers of the network.

ACKNOWLEDGMENTS

Expanse supercomputer time was provided via XSEDE allocations and NSG (the Neuroscience Gateway). We thank Amitava Majumdar, Subhashini Sivagnanam, and Kenneth Yoshimoto for providing computational resources.

We also thank professor Xiaolong Wang of UC San Diego for his teaching and guidance.